\UseRawInputEncoding

\documentclass[letterpaper, 10 pt, conference]{IEEEtran}  

\IEEEoverridecommandlockouts                              

\usepackage{amsmath,amssymb,amsfonts}
\usepackage{algorithmic}
\usepackage{algorithm}
\usepackage{array}
\usepackage[caption=false,font=normalsize,labelfont=sf,textfont=sf]{subfig}
\usepackage{textcomp}
\usepackage{stfloats}
\usepackage{url}
\usepackage{verbatim}
\usepackage{graphicx}
\usepackage{cite}
\usepackage{color}
\usepackage{makecell}
\title{\LARGE \bf
Stabilizing Dynamic Systems through Neural Network Learning: A Robust Approach
}

\author{Yu Zhang$^{1}$, Haoyu Zhang$^{1}$, Yongxiang Zou$^{1}$, Houcheng Li$^{1}$ and Long Cheng$^{1}$

\thanks{This work was supported in part by the National Natural Science Foundation of China under Grants 62025307, 62333023, and 62311530097.
	
	Y. Zhang, Y., Zou, H. Zhang,  X. Xia and L. Cheng, are all with the School of Artificial Intelligence, University of Chinese Academy of Sciences, Beijing 100049, China. They are also with the State Key laboratory of Multimodel Artificial Intelligence Systems, Institute of Automation, Chinese Academy of Sciences, Beijing 100190, China.  All correspondences should be addressed to the corresponding author Dr. Long Cheng (long.cheng@ia.ac.cn).}%
}

\begin{document}

\maketitle
\thispagestyle{empty}
\pagestyle{empty}
\columnsep 0.2in

\begin{abstract}

Point-to-point and periodic motions are ubiquitous in the world of robotics. To master these motions, Autonomous Dynamic System (DS) based algorithms are fundamental in the domain of Learning from Demonstration (LfD). However, these algorithms face the significant challenge of balancing precision in learning with the maintenance of system stability. This paper addresses this challenge by presenting a novel ADS algorithm that leverages neural network technology. The proposed algorithm is designed to distill essential knowledge from demonstration data, ensuring stability during the learning of both point-to-point and periodic motions. For point-to-point motions, a neural Lyapunov function is proposed to align with the provided demonstrations. In the case of periodic motions, the neural Lyapunov function is used with the transversal contraction to ensure that all generated motions converge to a stable limit cycle. The model  utilizes a streamlined neural network architecture, adept at achieving dual objectives: optimizing learning accuracy while maintaining global stability. To thoroughly assess the efficacy of the proposed algorithm,  rigorous evaluations are conducted using the LASA dataset and a manually designed dataset. These assessments were complemented by empirical validation through robotic experiments, providing robust evidence of the algorithm's performance
\end{abstract}

\section{INTRODUCTION}

Robotics is becoming more and more popular as technology gets better. In the future, robots are likely to be smarter and able to do complicated tasks comparable to those performed by people. However, programming robots to do things can be tough, especially for those who aren't experts. That's where Learning from Demonstration (LfD) comes in. LfD is an approachable method for robots to gain new skills. It allows them to learn by simply observing a person perform a task. This method eliminates the need for intricate programming or the use of complex reward mechanisms that are often required in traditional robot training. \cite{b1}.			

In practical applications, it's really important to make sure that the paths robots follow are stable. This helps to guarantee that they can smoothly reach the desired states. Dynamic Systems (DS) are sophisticated analytical tools that provide a flexible framework for modeling trajectories and for generating highly stable real-time motion control solutions \cite{b2}. 

A significant advantage of Autonomous DS lies in their capacity to encode the target states of a task into stable attractors, thereby endowing them with an intrinsic robustness to disturbances. Consequently, DS system can generate stable trajectories from any points to the attractors. This inherent stability not only enables seamless adaptation to novel conditions but also ensures sustained robust generalization performance. Furthermore, DS  are endowed with the capability to dynamically recalibrate a robot's trajectory in response to shifts in the target position or the emergence of unforeseen obstacles \cite{b3}. 

Considering the paramount importance of stability, ensuring its presence within DS is a critical aspect of their design. In the context of learning point-to-point motions, an elegant and robust approach to guaranteeing stability is introducing a positive-definite, continuously differentiable Lyapunov function, which acts as a reliable indicator of system stability.
The pioneering study that integrates Lyapunov theorems with DS learning is widely recognized as the Stable Estimator of Dynamical Systems  (SEDS) \cite{b2}.  This work has set a precedent in the field by demonstrating the efficacy of combining established stability analysis techniques with modern learning paradigms. In SEDS, the primary goal is to develop a DS that is globally asymptotically stable. This system is characterized by its representation through Gaussian Mixture Models (GMM) and Gaussian Mixture Regression (GMR). Crucially, the learning process in SEDS is designed to strictly conform to the constraints imposed by Lyapunov stability theory.

However, it is important to acknowledge that the application of a quadratic Lyapunov function within the SEDS may impose certain limitations on the accuracy of system reproduction. The academic community has increasingly recognized that the imposition of excessively stringent stability constraints can hinder the fidelity of learning from demonstrations. This, in turn, may result in less precise replications of the system's behavior \cite{b4}.

In response to the challenge of accuracy limitations in the SEDS, researchers have advanced a series of innovative modifications to the original approach. These adaptations aim to refine the reproduction accuracy of DS, thereby enhancing the overall fidelity of the learning process. \cite{b4,b5}. 

In contrast to traditional learning methods, a body of recent research has demonstrated that neural network-based algorithms \cite{b6,b7,b8,b9,b10,bzy} offer significant advantages in terms of LfD.  Neural networks can be conceptualized as versatile fitting functions capable of adapting to various data patterns. When structured or constrained appropriately, these networks are capable of generating trajectories that converge to meet the requirements of the Lyapunov function. The Lyapunov function itself can be either learned autonomously by a separate neural network or meticulously designed by human experts. In the ongoing quest to enhance the accuracy of system reproduction, some researchers have introduced specialized neural networks tailored for estimating the Lyapunov function from demonstrations \cite{b11}. This  method enhances the system's ability to accurately replicate observed data by addressing the limitations of the Lyapunov function. Despite these refinements, the foundational architecture of the neural network, as outlined in \cite{b11}, may require further optimization to achieve superior replication fidelity.

To address the limitations, recent research has advocated for the incorporation of invertible neural network architectures, as outlined in several studies (\cite{b6}, \cite{b7}, \cite{b8}, \cite{b9}, \cite{b10}, \cite{bzy}). These architectures are designed to convert the original DS into a more manageable form while preserving the critical attributes necessary for achieving convergence towards the target state. A key advantage of employing invertible transformations is the preservation of the stability inherent is the simplified model in the implicit space, which is essential for reliable long-term performance. Although this strategy can significantly boost the model's ability to capture intricate relationships, it's important to note that invertible neural networks typically have limited nonlinear expressiveness. To achieve optimal performance, it may be necessary to integrate multiple invertible neural networks. This approach, while effective, entails a trade-off: the computational complexity is increased, and the processing time is consequently extended.

To tackle the inherent challenges faced by algorithms reliant on invertible neural networks, this paper introduces an innovative approach that employs a directly neural network-based modeling of Lyapunov functions, thereby circumventing the need for invertible transformations. 

When learning periodic motions, a multitude of methodologies have been proposed for the modeling of periodic motion. In the seminal works within the fields of robotics and neuroscience, the use of limit cycles and central pattern generators (CPGs) has been pivotal for the simulation of locomotive periodic motions (\cite{b112,b113,b114,b115}). These early approaches laid the groundwork for understanding and replicating the complex patterns of movement observed in biological systems. However, formulations based on limit cycles have inherent limitations when it comes to learning complex periodic trajectories with traditional methods, which may be essential for several tasks. To overcome the limitations associated with traditional limit cycle formulations, this study introduces a novel algorithm grounded in neural network-based limit cycle learning. This algorithm is designed to adeptly acquire complex motion patterns while simultaneously ensuring the stability of the learned trajectories.

This paper makes several significant contributions to the field, which can be outlined as follows:
\begin{itemize}
	\item Two simply neural networks are used in this paper, one network is tasked with directly learning a Lyapunov function from the provided dataset. This function is instrumental in ensuring the stability of the system. Concurrently, the second network is designed to output velocities that are consistent with the derived Lyapunov function, thereby facilitating the generation of stable and coherent motion trajectories.	
	\item A minor modification is introduced to the proposed neural network, which endows it with the capability to learn a limit cycle characterized with a transverse contraction property.
	\item The experimental results provide conclusive evidence of the robustness and practicality of the proposed approach across a diverse range of learning scenarios.
\end{itemize}

The structure of this paper is delineated as follows: Section \ref{sec2} offers a thorough examination of the problem formulations, establishing a solid foundation for the subsequent discussions. Moving forward, Section \ref{sec3} delves into the intricate details of the neural networks employed in our study, elucidating their design and functionality. Section \ref{sec4} showcases the evaluation results of the proposed algorithm, supported by empirical data and referenced studies such as \cite{b12}, as well as experimental validations conducted on the Franka robot. Finally, Section \ref{sec5} encapsulates the conclusions drawn from the research, summarizing the key findings and their implications.

\section{Problem Formulation}\label{sec2}
Motion can typically be categorized into two fundamental types: point-to-point and periodic motions. These motions are often efficiently modeled using an autonomous DS, which can be represented as follows:
\begin{equation}\label{equ:1}
	\dot{\boldsymbol{x}} = f(\boldsymbol{x}), \ \forall \boldsymbol{x} \in \mathbb{R}^{d},
\end{equation} where $f: \mathbb{R}^{d} \mapsto \mathbb{R}^{d}$ is a continuous and continuously differentiable function.  This function is characterized by its properties relevant to the motion types: for point-to-point motions, it possesses a unique equilibrium state, while for periodic motions, it exhibits closely spaced equilibrium states.   Equation (\ref{equ:1}), which can also describe the manipulation skills, yields a solution $\Phi(\boldsymbol{x}_{0},t)$ that represents a valid trajectory generated when provided with  $\boldsymbol{x}_{0}$.  Consequently, by altering the initial conditions $\boldsymbol{x}_{0}$, different  trajectories leading to the target states can be generated. This flexibility allows for various trajectories to be produced, each tailored to different initial conditions.

When learning from demonstrations, it is presupposed that these demonstrations adhere to the characteristics of an Autonomous DS as defined by Equation (\ref{equ:1}). This assumption of alignment enables a parametric modeling approach for the system, which can be mathematically represented as follows:
\begin{equation}\label{equ:2}
	\hat{\dot{\boldsymbol{x}}} = \hat{f}(\boldsymbol{x},\boldsymbol{\theta}), \ \forall \boldsymbol{x} \in \mathbb{R}^{d}.
\end{equation}
where $\hat{\dot{\boldsymbol{x}}}$ represents the predicted value of the actual $\dot{\boldsymbol{x}}$, and $\hat{f}$ signifies a manually designed model intended to approximate the behavior of  the DS as depicted in (\ref{equ:1}). The model's parameters, denoted by $\boldsymbol{\theta}$  are the subject of the learning process. The optimal parameter configuration, represented as $\boldsymbol{\theta^{*}}$, is obtained by minimizing the following objective function:
\begin{equation}\label{equ:3}
	J(\boldsymbol{\theta} ) \propto \sum_{n=1}^{N_{d}}\sum_{k=1}^{K_{n}}\left \| \hat{\dot{\boldsymbol{x}}}_{k,n}-\dot{\boldsymbol{x}}_{k,n} \right \|^{2},
\end{equation}
where $n$ denotes the index of the demonstrations, $N_d$ signifies the total number of demonstrations available for learning,  $k$ represents the sampling time step, $K_n$ denotes the total number of sampling instances within the 
n-th demonstration. Meanwhile, $\propto$ indicates a proportionality relation, $\left \| \cdot  \right \|^{2}$ refers to the $l_{2}$-norm and $\hat{\dot{\boldsymbol{x}}}_{k,n}$ means the estimate of $\dot{\boldsymbol{x}}_{k,n}$. 
Hence, the objective function is constructed to minimize the difference between these predictions and the observed velocities, ensuring that the learned model accurately reproduces the demonstrated behavior.

When learning the point-to-point motion, it is imperative to recognize that the objective function alone cannot ensure the stability of the model. The stability of an Autonomous DS at an equilibrium point $\boldsymbol{x}^*$ is established by the existence of a Lyapunov candidate function $V(\boldsymbol{x}): \mathbb{R}^d\rightarrow \mathbb{R}$ that satisfies the following conditions:
\begin{equation}\label{equ:4}
	\begin{cases}
		(a)	V(\boldsymbol{x}^*)=0, \\
		(b)	\dot{V}(\boldsymbol{x}^*)=0, \\
		(c) V(\boldsymbol{x})>0 :\forall \boldsymbol{x}\neq \boldsymbol{x}^*, \\
		(d)	\dot{V}(\boldsymbol{x})<0 :\forall \boldsymbol{x}\neq \boldsymbol{x}^*.
	\end{cases}
\end{equation}
Furthermore, if the system satisfies the condition of radial unboundedness:
\begin{equation}\label{equ:5}
	\lim_{ \left \| \boldsymbol{x} \right \|\rightarrow +\infty }V(\boldsymbol{x})=+\infty,
\end{equation} the DS exhibits global asymptotic stability.

When learning periodic motion, it is crucial to understand that the objective function alone is also insufficient to guarantee the stability of the model. With equation (\ref{equ:1}), the following relation holds:
\begin{equation}\label{equ:6}
	\delta \dot{\boldsymbol{x}}(t) = \frac{\partial f(\boldsymbol{x}(t))}{\partial \boldsymbol{x}(t)} 	\delta \boldsymbol{x}(t),
\end{equation}
where $\delta \boldsymbol{\boldsymbol{x}}(t)$ denotes an infinitesimal virtual displacement at time $t$. The squared virtual displacement between two trajectories of equation (\ref{equ:1}) in a symmetric uniformly positive definite metric $M(\boldsymbol{x}(t)) \in \mathbb{R}^{n\times n}  $ is given by 
\begin{equation}\label{equ:7}
v(\boldsymbol{x}(t), \delta \boldsymbol{x}(t)) = \delta \boldsymbol{x}(t)^{^\mathrm{T}}   M(\boldsymbol{x}(t))     \delta \boldsymbol{x}(t)		
\end{equation}
As presented in \cite{}, if the following inquality is satisfied:
\begin{equation}\label{equ:8}
\frac{\partial v(\boldsymbol{x}, \delta \boldsymbol{x})}{\partial \boldsymbol{x}} f(\boldsymbol{x})+\frac{\partial v(\boldsymbol{x}, \delta \boldsymbol{x})}{\partial \delta \boldsymbol{x}} \frac{\partial f(\boldsymbol{x})}{\partial  \boldsymbol{x}}\delta \boldsymbol{x} \le  -\varsigma  v(\boldsymbol{x}, \delta \boldsymbol{x}) ,
\end{equation}
for all $\delta \boldsymbol{x} \ne 0$ and $\lambda > 0$, the system defined in (\ref{equ:1}) is said to be transverse contracting, and all solutions starting in the $strictly$ $forward$ $invariant$ set converge to a unique limit cycle.

Incorporating stability as a pivotal element in the DS learning process is crucial for ensuring that a robot reliably attains its intended target states. However, integrating stability considerations can sometimes inadvertently constrain the model's precision.

To improve precision,  a data-driven methodology is introduced that harnesses the capabilities of neural networks to identify the Lyapunov function. This innovative strategy bypasses the conventional need for manually designing a Lyapunov candidate function, thereby simplifying the process of ensuring stability within the DS learning framework. The following section delves into the integration of this method into the DS learning process, illustrating its effectiveness in meeting the desired stability criteria in both point-to-point learning and limit cycle learning scenarios.

\section{The Proposed Approach}\label{sec3}

For learning point-to-point motions, ensuring system stability is commonly achieved by incorporating a Lyapunov function,  without loss of generality, the motion targets are positioned at the origin of the Cartesian coordinate system and in this paper,  the Lyapunov function is established as:
\begin{equation}\label{equ:9}
	V(\boldsymbol{x})=(g(\boldsymbol{x})+\delta) (\boldsymbol{x}^\mathrm{T}\boldsymbol{x}),
\end{equation}
where $g(\cdot)$ represents an arbitrary neural network but using a $softplus$ or $sigmoid$ activation function after the final layer to guarantee non-negative across all possible inputs. Meanwhile, $\delta$ denotes a small positive constant, therefore, the condition $c$ as outlined in conditions (\ref{equ:4}) is satisfied. Additionally,  the term $(\boldsymbol{x}^\mathrm{T}\boldsymbol{x})$ serves to satisfy the condition $a$ and $b$.

Ensuring the system's stability necessitates the computation of the Lyapunov energy function's derivative, which can be formulated as:
\begin{equation}\label{equ:10}
	\dot{V}(\boldsymbol{x})=\frac{\partial V}{\partial \boldsymbol{x}} \boldsymbol{\dot{x}}.
\end{equation}
Utilizing the automatic differentiation of "PyTorch" framework,  $\frac{\partial V}{\partial \boldsymbol{x}}$ can be directly acquired.  
To satisfy the condition $d$ in conditions (\ref{equ:4}),  
 $\dot{\boldsymbol{x}}$ is designed as:
\begin{equation}\label{equ:11}
	\dot{\boldsymbol{x}}=-\alpha(\boldsymbol{x})(\frac{\partial  V}{\partial \boldsymbol{x}} f(\boldsymbol{x}))f(\boldsymbol{x})-\beta(\boldsymbol{x}) f(\boldsymbol{x})/(\frac{\partial  V}{\partial \boldsymbol{x}} f(\boldsymbol{x})), 
\end{equation}
where $f(\cdot)$ denotes an arbitrary neural network, $\alpha(\boldsymbol{x})$ and $\beta(\boldsymbol{x})$  represent positive scaling factors. They are all adaptively learned from the training dataset.

Since $\frac{\partial  v}{\partial \boldsymbol{x}} f(\boldsymbol{x})$ is a scalar, and $\dot{V}(\boldsymbol{x})$ can be calculated as:
\begin{equation}\label{equ:12}
		\dot{V}(\boldsymbol{x})=-\alpha(\boldsymbol{x})(\frac{\partial  V}{\partial \boldsymbol{x}} f(\boldsymbol{x}))^2-\beta(\boldsymbol{x}), 
\end{equation}
therefore, the condition $d$ in (\ref{equ:4}) is also satisfied and the system is stable.

Typically, without prioritizing stability, $f(\cdot)$ can
effectively learn from demonstrations. This implies that retaining more features from
$f(\cdot)$ while simultaneously ensuring stability through the proposed algorithm may lead to enhanced performance. 

To keep more features from $f(\cdot)$, the $f(\cdot)$ is first decomposed into two parts as:
$\boldsymbol{R}_1f(\cdot)$ and $\boldsymbol{R}_2f(\cdot)$, where $\boldsymbol{R}_1$ and $\boldsymbol{R}_2$ are  projection matrices as:
\begin{equation}\label{equ:13}
	\begin{aligned}
		\begin{cases}
			
		\boldsymbol{R}_1&=\boldsymbol{I}-\frac{(\frac{\partial  V}{\partial \boldsymbol{x}})^\mathrm{T}\frac{\partial  V}{\partial \boldsymbol{x}}}{\frac{\partial  V}{\partial \boldsymbol{x}}(\frac{\partial  V}{\partial \boldsymbol{x}})^\mathrm{T}} \\
		\boldsymbol{R}_2&=\frac{(\frac{\partial  V}{\partial \boldsymbol{x}})^\mathrm{T}\frac{\partial  V}{\partial \boldsymbol{x}}}{\frac{\partial  V}{\partial \boldsymbol{x}}(\frac{\partial  V}{\partial \boldsymbol{x}})^\mathrm{T}}	
\end{cases}	
	\end{aligned}
\end{equation}
Therefore, the first part $f_1(\boldsymbol{x})=\boldsymbol{R}f(\boldsymbol{x})$ does not influence the system's stability as the equation $\frac{\partial  V}{\partial \boldsymbol{x}}\boldsymbol{R}_1=\boldsymbol{0}$ always holds. 

The second part $f_2(\boldsymbol{x})$ is processed to guarantee the stability by using the proposed algorithm as:
\begin{equation}\label{equ:14}
	f_2(\boldsymbol{x})=-\alpha(\boldsymbol{x})(\frac{\partial  V}{\partial \boldsymbol{x}}  f(\boldsymbol{x}))\boldsymbol{R}_2f(\boldsymbol{x})-\beta(\boldsymbol{x})\boldsymbol{R}_2 f(\boldsymbol{x})/(\frac{\partial  V}{\partial \boldsymbol{x}} f(\boldsymbol{x}))
\end{equation}
and finally, the output of the model is designed as:
\begin{equation}\label{equ:15} \dot{\boldsymbol{x}}=f_1(\boldsymbol{x})+f_2(\boldsymbol{x}).
\end{equation}

In real-life scenarios, point-to-point motion models are capable of capturing the majority of motions, however, periodic motions are also frequently observed. By making appropriate modifications to equation (\ref{equ:14}), it is possible to learn and model these periodic motions. 

 In the context of learning periodic motions, the limit cycle serves as a robust tool to ensure the stability of the generated motions. While limit cycles are inherently suited for two-dimensional motions, higher-dimensional motions can be approached by composing multiple limit cycles when the dimension is an even number. In the case of an odd number of dimensions, the additional dimension can be modeled as a function of the other dimensions.

 For point-to-point motions, the proposed algorithm ensures that the energy of the Lyapunov function asymptotically approaches zero, indicative of the system's convergence to a fixed point. In contrast, when learning a limit cycle, the objective is to converging towards a closed trajectory. In this context it requires a nuanced expectation: the function should maintain a strictly positive value everywhere except along the cycle, where it should be zero. 

Considering the Lyapunov function as:
\begin{equation}\label{equ:16}
	V(\boldsymbol{x})=\frac{1}{2} (g(\boldsymbol{x})-\delta)^2 ,
\end{equation}
where $g(\cdot)$ and  $\delta$ are designed as in equation (\ref{}). 
  The derivation of the Lyapunov function can be calculated as:
\begin{equation}\label{equ:17}
	\dot{V}(\boldsymbol{x})=(g(\boldsymbol{x})-\delta) \frac{\partial g(\boldsymbol{x})}{\partial \boldsymbol{x}} \boldsymbol{\dot{x}},
\end{equation}

Therefore, as the proposed algorithm, $\dot{x}$ can be designed as:
\begin{equation}\label{equ:18}
		\begin{aligned}
	\dot{\boldsymbol{x}}= &-\alpha(\boldsymbol{x})(g(\boldsymbol{x})-\delta)(\frac{\partial  g(\boldsymbol{x})}{\partial \boldsymbol{x}} f(\boldsymbol{x}))f(\boldsymbol{x})  \\	
	&-\beta(\boldsymbol{x})(g(\boldsymbol{x})-\delta) f(\boldsymbol{x})/(\frac{\partial  g(\boldsymbol{x})}{\partial \boldsymbol{x}} f(\boldsymbol{x})), 
		\end{aligned}
\end{equation}

A preliminary analysis of the equation (\ref{equ:15}) reveals that learned DS can be decomposed into two components that govern the motion. The first component is designed to capture the essential features of the demonstrations, while the second component ensures the stability of the system.
When learning the limit cycle, the projection matrices are used again as:

\begin{equation}\label{equ:19}
	\begin{aligned}
		\begin{cases}			
			\boldsymbol{R}_3&=\boldsymbol{I}-\frac{(\frac{\partial  g(\boldsymbol{x})}{\partial \boldsymbol{x}})^\mathrm{T}\frac{\partial  g(\boldsymbol{x})}{\partial \boldsymbol{x}}}{\frac{\partial  g(\boldsymbol{x})}{\partial \boldsymbol{x}}(\frac{\partial  g(\boldsymbol{x})}{\partial \boldsymbol{x}})^\mathrm{T}} \\
			\boldsymbol{R}_4&=\frac{(\frac{\partial  g(\boldsymbol{x})}{\partial \boldsymbol{x}})^\mathrm{T}\frac{\partial  g(\boldsymbol{x})}{\partial \boldsymbol{x}}}{\frac{\partial  g(\boldsymbol{x})}{\partial \boldsymbol{x}}(\frac{\partial  g(\boldsymbol{x})}{\partial \boldsymbol{x}})^\mathrm{T}}	
		\end{cases}	
	\end{aligned}
\end{equation}

The first term of the equation, as  (\ref{equ:14}), exerts no influence on the system's stability, thus can be used in learning of the limit cycle as $f_3(\boldsymbol{x}) = \boldsymbol{R}_3 f(\boldsymbol{x})$. The second part is used to guarantee the stability and is modified as:
\begin{equation}\label{equ:20}
	\begin{aligned}
	f_4(\boldsymbol{x})= &-\alpha(\boldsymbol{x})(g(\boldsymbol{x})-\delta)(\frac{\partial  g(\boldsymbol{x})}{\partial \boldsymbol{x}} f(\boldsymbol{x}))\boldsymbol{R}_4f(\boldsymbol{x})  \\	
		&-\beta(\boldsymbol{x})(g(\boldsymbol{x})-\delta) \boldsymbol{R}_4f(\boldsymbol{x})/(\frac{\partial  g(\boldsymbol{x})}{\partial \boldsymbol{x}} f(\boldsymbol{x})). 
	\end{aligned}
\end{equation}
Then the output of the model can be designed as:
\begin{equation}\label{equ:21} \dot{\boldsymbol{x}}=f_3(\boldsymbol{x})+f_4(\boldsymbol{x}),
\end{equation}
for learning the limit cycle. However, the method does not guarantee that only one limit cycle exist, to satisfy this constraint, the transverse contracting criteria is introduced.

Firstly, the $M(\boldsymbol{x}(t))$ is (\ref{equ:7}) is designed as: $M(\boldsymbol{x}(t))=g(\boldsymbol{x})\boldsymbol{I}$,
to satisfy (\ref{equ:8}), the inequality can be equivalent to:

\begin{equation}\label{equ:22}
	\frac{\partial g(\boldsymbol{x})}{\partial \boldsymbol{x}} \boldsymbol{\dot{x}}+\lambda_{max}   g(\boldsymbol{x})  -\varsigma g(\boldsymbol{x}) \le 0 ,
\end{equation}
where $\lambda_{max}$ is the max eigenvalue of $((\frac{\partial \boldsymbol{\dot{x}}}{\partial \boldsymbol{x}})^\mathrm{T}+\frac{\partial \boldsymbol{\dot{x}}}{\partial \boldsymbol{x}})$. Rewriting $\frac{\partial g(\boldsymbol{x})}{\partial \boldsymbol{x}} \boldsymbol{\dot{x}}+\lambda_{max}   g(\boldsymbol{x})  -\varsigma g(\boldsymbol{x})$ as $T(\boldsymbol{x})$, the output of the model to learn the limit cycle is as:
\begin{equation}\label{equ:21} \dot{\boldsymbol{x}}=f_3(\boldsymbol{x})+f_4(\boldsymbol{x})-(\xi f_3(\boldsymbol{x})+f_4(\boldsymbol{x})) \mathrm{Relu}(T(\boldsymbol{x}))/T(\boldsymbol{x}),
\end{equation}
where $\xi$ is a positive constant  close to one, $\mathrm{Relu}$ denotes the element-wise calculated function as:
\begin{equation}\label{equ:14}
	\mathrm{Relu}(s) = 	\begin{cases}
		s \ \ \ \ \ \ \ \ \  \mathrm{if} \ s>0, \\
		0 \ \ \ \ \ \ \ \ \	\mathrm{otherwise}.
	\end{cases}
\end{equation}

Consequently, the conditions for transverse contraction are met, enabling the acquisition of the limit cycle.

\begin{figure*}[htbp]
	\centerline{\includegraphics[width=1\textwidth]{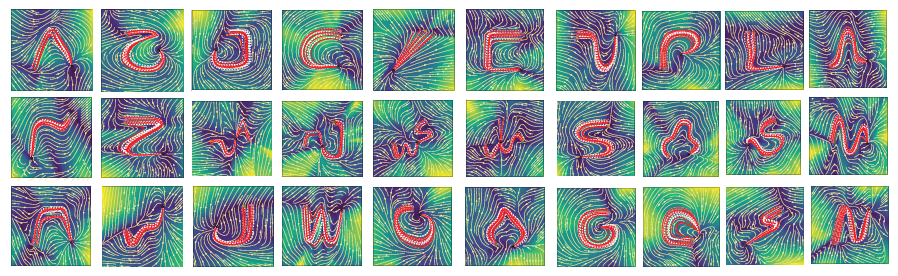}}
	\caption{
		The simulation utilizing the proposed algorithm is depicted in this paper. Images with a black-yellow background display the learned vector fields. Within these visuals, the white dotted lines represent the original demonstration data, while the red solid lines depict the reproductions from identical initial points. The target points are denoted as "$\cdot$" in these illustrations.	}	
	\label{fig2}
\end{figure*}

\section{ Experiment Results and Discussions}\label{sec4}
To thoroughly evaluate the performance of the proposed algorithm, a stringent assessment framework has been executed, integrating both simulated environments and empirical trials. The findings are delineated as follows:

\subsection{ Simulation Results}
Firstly, a simulation was carried out using the LASA dataset. 

To assess the effectiveness of the proposed algorithm, the quantitative evaluation of accuracy was performed through the application of two error metrics: the Swept Error Area (SEA) \cite{b5} and the Velocity Root Mean Square Error ($V_{rmse}$) \cite{b14}. The SEA scores serve as an indicator of the algorithm's capability to faithfully replicate the shapes of motions. On the other hand, the $V_{rmse}$ metric gauges the algorithm's proficiency in preserving the velocities inherent in demonstrations. A lower SEA signifies superior accuracy in reproducing the trajectories, while a lower $V_{rmse}$ indicates that the reproductions closely emulate the smoothness observed in the original demonstrations. These two metrics comprehensively evaluate the performance of the reproductions by quantifying their resemblance to the demonstrated trajectories.

For the implementation of the proposed algorithm, the "AdamW" optimization method is opted. Specifically, the learning rate was configured at $1 \times 10^{-5}$,   coupled with a decay rate of $0.99$. Subsequently, the trained DS model was employed to generate trajectories starting from identical points and with matching step increments as the original demonstrations. In this paper, all  data from the LASA dataset were utilized, with trajectories generated from all available starting points. Prior to processing, the data underwent normalization, bringing it within the range of [$-1, 1$]. The algorithm's parameters were initialized with random values, and the maximum number of iterations was set to 2000, with a mini-batch size of 64. For the sake of equitable comparisons,  DS algorithms that named SEDS and CLF-DM are chosen, for which source code was readily accessible. The parameters for these comparative algorithms were configured to match the values originally specified in their respective references, ensuring a fair benchmark.

Fig. \ref{fig2} presents a visual representation of the vector fields (depicted with dark background) and the transformed trajectories generated by the proposed DS algorithm for 29 examples drawn from the LASA dataset. Notably, the reproduced trajectories (depicted in red) closely align with the original demonstrations (depicted in white). It is worth emphasizing that  irrespective of whether the starting points align with those of the demonstrations, the reproduced DS consistently converges toward the intended goal. Furthermore, the last three images in the third row and first image in the fourth row of Fig. \ref{fig2} illustrate more intricate motions. For instance, in the third image of the third row, three different types of demonstrations commence from distinct initial points but all culminate at the same target point, underscoring the algorithm's versatility and effectiveness.

The quantified results are shown Table \ref{tab:table2}, the results clearly demonstrate that the proposed algorithm outperforms other methods in terms of trajectory reproduction accuracy. Showcasing substantial improvements of approximately $24.6\%$ in SEA score and approximately $16.1\%$ in $V_{rmse}$ when compared to CLF-DM.
\begin{figure}[htbp]
	\centerline{\includegraphics[width=0.6\textwidth]{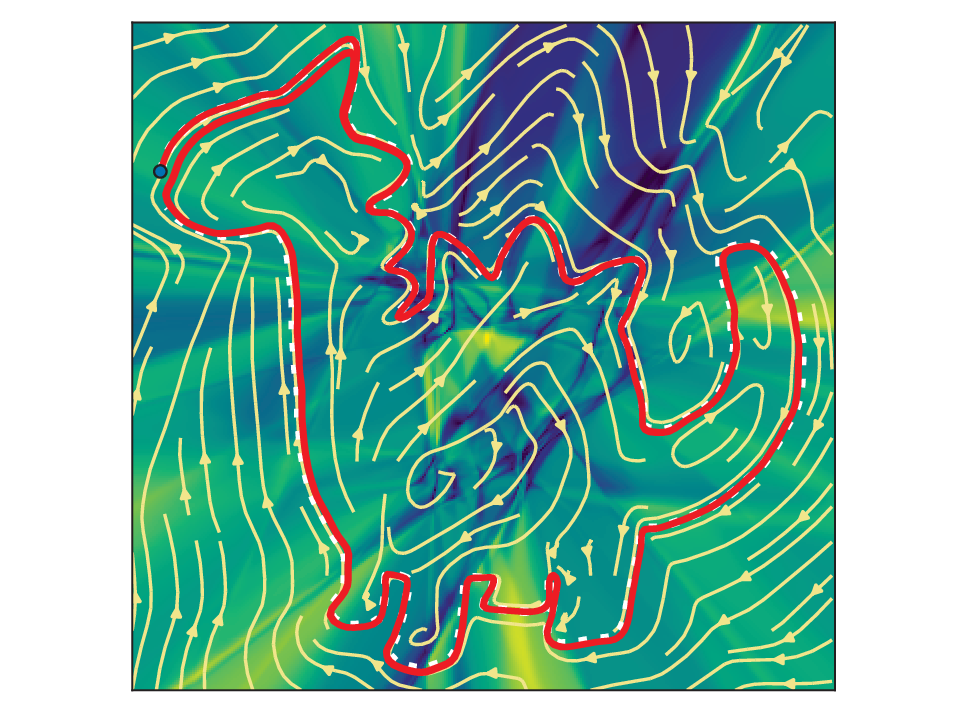}}
	\caption{The proposed algorithm to learn a complex periodic trajectory, the white dotted lines represent the original demonstration data, while the red solid lines depict the reproductions from identical initial points.  }	
	\label{fig2}
\end{figure}

\begin{table}[t]
	\caption{Variation in Reproduction Errors of different DS approaches on the LASA dataset\label{tab:table2}}
	\centering
	\setlength{\tabcolsep}{1.4mm}{	\begin{tabular}{ccc}
			\hline
			 methods    & Mean SEA($mm^2$)   & Mean $V_{rmse}(mm/s)$
			\\
			\hline
			\makecell*[c]{SEDS \cite{b2}} & 8.18 $\times$ 10\textasciicircum{}5 & 142.9  \\
			
			\makecell*[c]{CLF-DM \cite{b5} } & 5.72 $\times$ 10\textasciicircum{}5 & 71.8  \\			
			
			\makecell*[c]{The proposed method  }  & 4.31 $\times$ 10\textasciicircum{}5 & 60.2  \\
			\hline		
	\end{tabular}}
\end{table}

The proposed algorithm is capable of effectively learning the limit cycle, a task that involves mastering a complex, manually designed periodic trajectory. This capability is illustrated in Figure \ref{fig2}, where the learned limit cycle is compared alongside the manually designed trajectory. The outcomes demonstrate the algorithm's proficiency in acquiring intricate periodic paths.

\begin{figure}[htbp]
	\centerline{\includegraphics[width=0.5\textwidth]{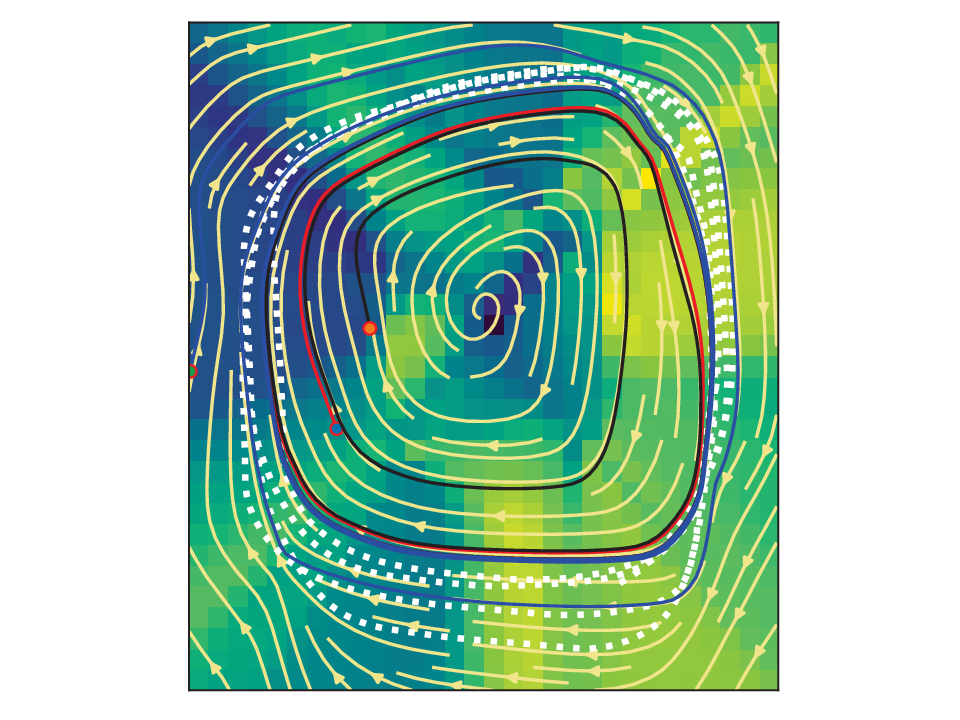}}
	\caption{The proposed algorithm is used to learn a limit cycle from actual trajectory data collected by a robot. In the visual representation, the white dotted lines correspond to the original demonstration data, whereas the colored solid lines illustrate the algorithm's generated trajectories initiated from various random starting points. }	
	\label{fig3}
\end{figure}

A demonstration trajectory, collected by the Franka robot, serves as the basis for learning the limit cycle. The outcome of this learning process is depicted in Fig. \ref{fig3}.

\begin{figure}[htbp]
	\centerline{\includegraphics[width=0.5\textwidth]{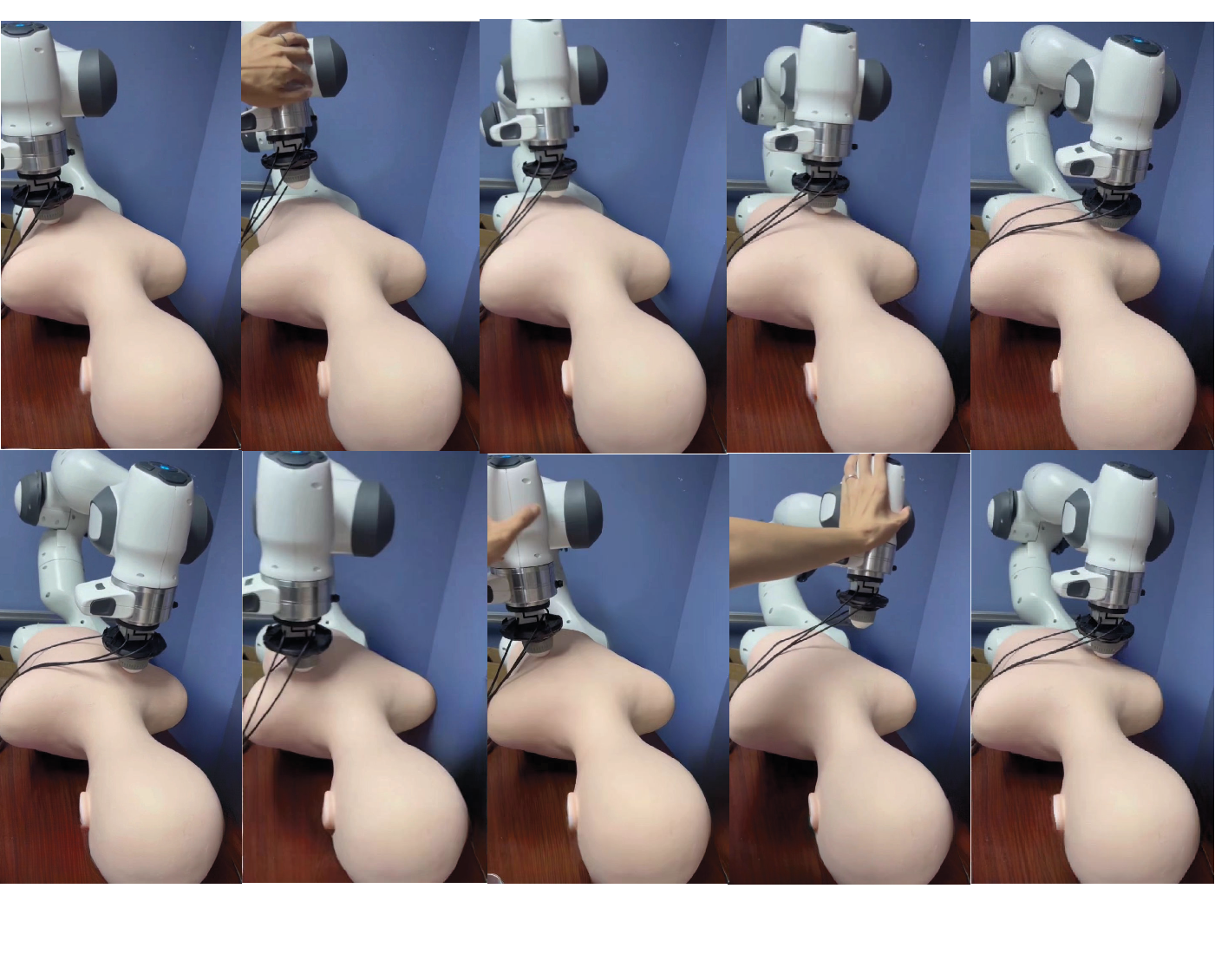}}
	\caption{The proposed algorithm is designed to perform a massage task, leveraging the learned limit cycle derived from the actual trajectory data collected from a robot. Despite encountering various disturbances during the massage process, the robot is capable of maintaining task performance effectively.  }	
	\label{fig4}
\end{figure}
\subsection{ Validation on Robot}
Experiments were conducted to validate the proposed algorithm using a Franka Emika robot. The model was trained using the collected data as in Fig. \ref{fig3}.

Throughout the experiment, the robot continuously acquired its end-effector's position and utilized the proposed algorithm to calculate the corresponding velocity. The impedance control is used and when in the real experiments, disturbance are forced by human to evaluate the robust of the proposed algorithm. 
Snapshots of the motions are shown in  Fig. \ref{fig4}.

\section{CONCLUSIONS}\label{sec5}

The present study introduces a neural network-based algorithm, tailored for learning from both point-to-point and periodic motions. The efficacy of this algorithm is rigorously evaluated through simulated scenarios encompassing a variety of handwriting samples, as well as practical experiments utilizing a physical robot. The experimental results offer substantial evidence supporting the effectiveness of the proposed methodology.

However, it is imperative to recognize that the algorithm's reliance on neural networks entails a relatively more time-intensive training process compared to traditional DMPs or GMMs. This consideration highlights an area for future research, focusing on improving the algorithm's computational efficiency, generalization capabilities, and reducing the duration of the training phase

\addtolength{\textheight}{-12cm}   





\end{document}